\documentclass[pdflatex,sn-mathphys-num]{sn-jnl}

\usepackage{graphicx}%
\usepackage{multirow}%
\usepackage{amsmath,amssymb,amsfonts}%
\usepackage{amsthm}%
\usepackage{mathrsfs}%
\usepackage[title]{appendix}%
\usepackage{xcolor}%
\usepackage{textcomp}%
\usepackage{manyfoot}%
\usepackage{booktabs}%
\usepackage{algorithm}%
\usepackage{algorithmicx}%
\usepackage{algpseudocode}%
\usepackage{listings}%

\usepackage{newfloat}

\usepackage{caption}
\usepackage{pifont}
\usepackage{subcaption}
\usepackage{mwe}
\usepackage{float} 
\usepackage{mathrsfs}

\newcommand{\cmark}{\ding{51}}%

\newcommand{\INVISIBLE}[1]{{}}

\theoremstyle{thmstyleone}%
\theoremstyle{thmstyletwo}%

\theoremstyle{thmstylethree}%

\let\orcidlogo\relax
\usepackage{orcidlink}

\begin{document}

\title[Article Title]{Tailoring Adverse Event Prediction in Type 1 Diabetes with Patient-Specific Deep Learning Models}

\author*[1]{\fnm{Giorgia} \sur{Rigamonti}\textsuperscript{\orcidlink{0009-0006-4253-1020}}}\email{giorgia.rigamonti@unimib.it}

\author[1]{\fnm{Mirko Paolo} \sur{Barbato}\textsuperscript{\orcidlink{0000-0002-3967-8957}}}\email{mirko.barbato@unimib.it}

\author[1]{\fnm{Davide} \sur{Marelli}\textsuperscript{\orcidlink{0000-0003-2389-0301}}}\email{davide.marelli@unimib.it}

\author[1]{\fnm{Paolo} \sur{Napoletano}\textsuperscript{\orcidlink{0000-0001-9112-0574}}}\email{paolo.napoletano@unimib.it}

\affil[1]{\orgdiv{Department of Informatics, Systems and Communication}, \orgname{University of Milano-Bicocca}, \city{Milano}, \postcode{20126}, \state{Italy}}


\abstract{Effective management of Type 1 Diabetes requires continuous glucose monitoring and precise insulin adjustments to prevent hyperglycemia and hypoglycemia. With the growing adoption of wearable glucose monitors and mobile health applications, accurate blood glucose prediction is essential for enhancing automated insulin delivery and decision-support systems. This paper presents a deep learning-based approach for personalized blood glucose prediction, leveraging patient-specific data to improve prediction accuracy and responsiveness in real-world scenarios.
Unlike traditional generalized models, our method accounts for individual variability, enabling more effective subject-specific predictions. We compare Leave-One-Subject-Out Cross-Validation with a fine-tuning strategy to evaluate their ability to model patient-specific dynamics. Results show that personalized models significantly improve the prediction of adverse events, enabling more precise and timely interventions in real-world scenarios.
To assess the impact of patient-specific data, we conduct experiments comparing a multimodal, patient-specific approach against traditional CGM-only methods.  
Additionally, we perform an ablation study to investigate model performance with progressively smaller training sets, identifying the minimum data required for effective personalization—an essential consideration for real-world applications where extensive data collection is often challenging. Our findings underscore the potential of adaptive, personalized glucose prediction models for advancing next-generation diabetes management, particularly in wearable and mobile health platforms, enhancing consumer-oriented diabetes care solutions.
}




\maketitle

\section{Introduction}\label{sec1}
Diabetic patients suffer from impaired insulin secretion or action, significantly affecting their quality of life~\cite{insulin}. Type 1 Diabetes (T1D), in particular, is a life-threatening condition that necessitates continuous external intervention~\cite{type_1}. To prevent dangerous fluctuations in blood glucose levels, individuals with diabetes must consistently monitor and regulate their insulin intake~\cite{Mathew2023}. 
In recent years, consumer electronics—such as wearable continuous glucose monitors, connected insulin pumps, and smartphone-based health applications—have played an increasingly critical role in diabetes self-management, enabling real-time data collection, monitoring, and intervention. Hyperglycemia, defined as a Blood Glucose Concentration (BGC) above 180 mg/dL~\cite{hyper}, and hypoglycemia, with BGC below 70 mg/dL~\cite{hypoglycemia}, pose serious health risks if not properly managed. Technological advancements have enabled Continuous Glucose Monitoring (CGM)~\cite{cgm}, which can be integrated with dynamic insulin pump therapy~\cite{csii} to deliver subcutaneous insulin and help maintain safe glucose levels. 

Accurately estimating future BGC is essential for adjusting insulin doses and avoiding adverse glycemic events. However, prediction remains highly challenging due to physiological variability and dynamic influences such as food intake, physical activity, stress, and sleep~\cite{bent2021engineering}.
Many recent approaches rely on a single, generalized model applied uniformly across individuals, overlooking the substantial inter-patient variability in glucose metabolism, insulin sensitivity, and lifestyle factors~\cite{6157604}. While some methods incorporate demographic features such as age and gender~\cite{deep}, these static characteristics often fall short in capturing the dynamic, time-varying nature of BGC~\cite{oviedo2017review}.
Real-world personalization remains limited, especially in T1D, where glucose regulation relies mainly on external interventions. Existing models are often complex, with high parameter counts that hinder adaptability, particularly when subject-specific data is limited~\cite{aliberti2019multi}. Moreover, many studies focus on Type 2 diabetes (T2D) and rely solely on CGM data~\cite{deng2021deep}, without modeling the critical interactions among insulin dosing, meal intake, and glucose levels that are central to T1D management~\cite{podobnik2025metabolic}.
To address these limitations, we argue that effective T1D prediction requires a multimodal, patient-specific approach—one that moves beyond CGM-only inputs and static personalization to capture individual metabolic responses more accurately.

Our analysis reveals that the variability in these logs is highly distinctive and can be used to classify patient identity with high confidence. This finding suggests that personalized deep learning models can significantly enhance prediction accuracy compared to one-size-fits-all approaches. 
To highlight the value of multimodal patient-specific data, we conduct comparative experiments with CGM-only predictions, which alone cannot describe a patient as well as other modalities. 

\noindent The key contributions of this work are summarized as follows:

\begin{itemize}
\item We demonstrate that individuals with T1D can be reliably identified based on their multimodal daily log data, highlighting strong inter-individual variability in physiological and behavioral glucose dynamics.
    
\item We propose a patient-specific deep learning approach that extends a Bi-GRU architecture through a two-phase process: (i) a patient-independent model trained using Leave-One-Subject-Out Cross-Validation (LOSOCV), and (ii) a fine-tuned model adapted to individual characteristics using limited subject-specific data.
    
\item We compare our approach to a state-of-the-art CGM-only baseline, showing that incorporating multimodal physiological data significantly improves accuracy, especially for detecting adverse glycemic events.
    
\item We perform an ablation study to evaluate model robustness under decreasing amounts of training data, identifying the minimum data required for effective personalization—an essential analysis for real-world applications.
\end{itemize}

\noindent These contributions enable the development of accurate, lightweight, and adaptive glucose prediction models that can be integrated into next-generation wearable devices, mobile health applications, and real-time consumer health technologies for personalized diabetes management.

\section{Literature Review}\label{sec1-related}
Machine Learning (ML) and Deep Learning (DL) techniques have significantly advanced BGC prediction, offering more flexible alternatives to traditional statistical models such as ARIMA~\cite{arima} and Unobserved Components Models (UCM)~\cite{blood}. While these newer approaches often outperform classical methods, they still struggle to fully capture the complex temporal dynamics and physiological variability inherent in real-world patient data.
In parallel with architectural innovation, multimodal approaches have emerged to integrate contextual and behavioral data. Models such as BG-BERT~\cite{bg-bert} and Bi-GRU~\cite{rigamonti2024improving} incorporate diverse input streams—including insulin administration, carbohydrate intake, and physical activity—resulting in more informative and clinically relevant predictions. However, most of these models remain generalized, applying a single architecture across heterogeneous patient populations. This often overlooks substantial inter-individual differences in glucose metabolism, insulin sensitivity, lifestyle, and treatment patterns. Furthermore, high model complexity can hinder applicability in settings with sparse or incomplete patient data~\cite{aliberti2019multi}.
To overcome these limitations, recent research has increasingly focused on personalized forecasting, tailoring models to individual patient profiles. Some approaches incorporate static features such as age and gender~\cite{deep}, but BGC dynamics are primarily influenced by time-varying factors including insulin dosing, diet, activity, stress, and sleep~\cite{oviedo2017review}, which must be captured in real time for accurate forecasting.
Most personalization efforts to date have concentrated on T2D, typically using CGM-only data~\cite{deng2021deep}, with limited extension to type 1 diabetes. This represents a critical gap: T1D management relies entirely on exogenous insulin, requiring models to explicitly account for the complex interactions between insulin administration, carbohydrate intake, and physiological response~\cite{podobnik2025metabolic}.
Effective forecasting for T1D thus demands multimodal, patient-specific models. Population-trained models often fail to generalize across individuals, while personalized strategies—such as fine-tuning~\cite{seo2021personalized,deng2021deep} and meta-learning~\cite{langarica2023meta,9813400}—have shown strong performance even with limited subject-specific data. Recent studies~\cite{langarica2024deep,lara2025personalized} confirm the practical value of such approaches in real-world settings.
In summary, inter-individual variability is not residual noise but a central challenge in BGC prediction. Future efforts—particularly for T1D—must prioritize adaptive, fine-tuned architectures that respond dynamically to each patient's unique metabolic patterns and daily fluctuations. Yet, few existing models explicitly address the dual challenge of inter-patient generalization and intra-patient personalization under multimodal, data-constrained conditions.

\section{Method}
 
In this section, we present a personalized methodology for BGC prediction in T1D subjects, structured as a two-stage learning strategy that accounts for both inter- and intra-individual variability using a Bi-GRU-based architecture. The approach comprises:
\begin{itemize}
    \item {Patient Identification}---We show that individual patients can be accurately identified from multimodal data (CGM, insulin administration, and carbohydrate intake) by casting the task as a classification problem, revealing distinct subject-specific glucose dynamics.  
    \item {Patient-Specific Prediction}---We use a two-phase DL pipeline: a patient-independent model is trained with LOSOCV to capture general dynamics, then fine-tuned on limited patient data for personalized glucose prediction.
\end{itemize}
This approach combines robust generalization with personalization, offering a practical and lightweight solution for real-world glucose prediction in consumer health applications. The classification task serves as a preliminary step to highlight subject-specific patterns later exploited in the patient-specific regression model.


\subsection{Patient Identification via Multimodal Data} 

\begin{figure*}[tb]
    \centering
    \includegraphics[width=1.0\textwidth]{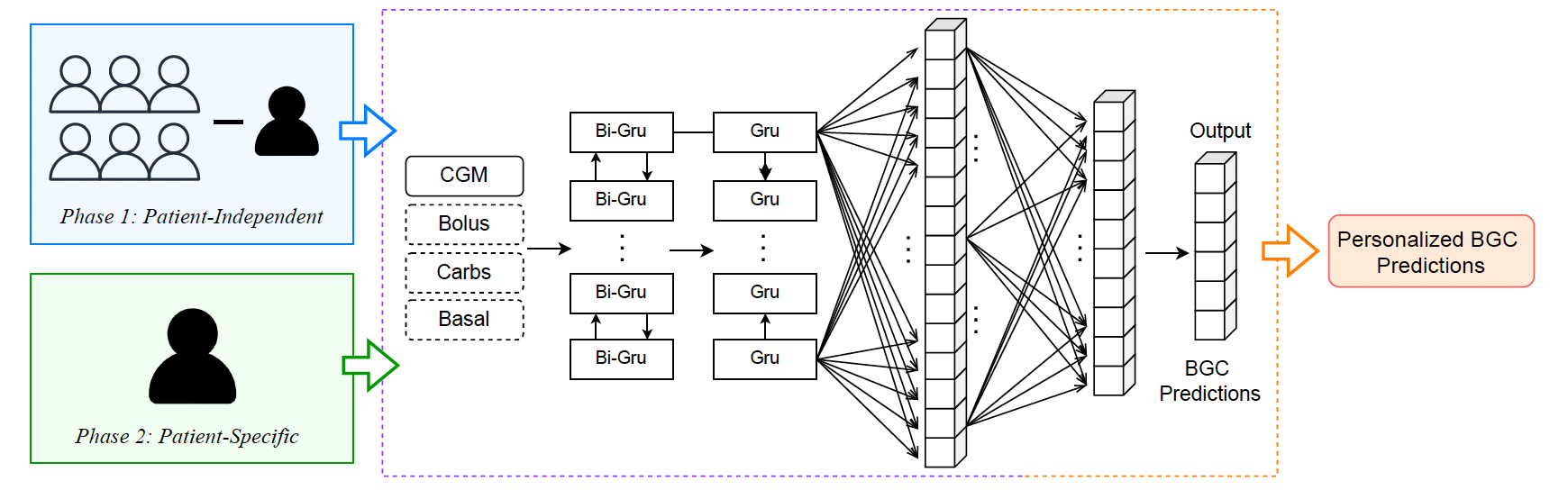}
    \captionsetup{font=small}
    \caption{Bi-GRU architecture combining patient-independent learning with patient-specific fine-tuning to produce personalized BGC predictions.}
    \label{fig:regression}
\end{figure*}

We begin by investigating the variability in T1D patient profiles and analyzing how different input features influence individual glucose trends. To this end, we propose a custom-designed architecture based on a Bidirectional Gated Recurrent Unit (Bi-GRU) network, tailored for the classification task of subject identification. This architecture is optimized to balance performance and computational efficiency, making it well-suited for learning patient-specific temporal patterns from multimodal glucose-related data. The model adopts an architecture analogous to the one depicted in Figure~\ref{fig:regression}, which will be later discussed in the context of the regression task, and differs only in the final layer, which, in this case, is designed for patient identification.
The proposed RNN consists of a Bi-GRU layer with 128 units, followed by a unidirectional GRU layer with 256 units. The Bi-GRU layer enables the model to extract temporal dependencies in both forward and backward directions, enhancing its capacity to learn contextual patterns. The GRU layer then further refines these representations, leveraging its efficiency in handling sequential physiological data.
To mitigate overfitting and promote generalization, we include a dropout layer with a rate of 0.4 after the recurrent layers, followed by a Fully Connected layer and a second dropout layer with the same rate. An average pooling layer is then applied to condense temporal information and reduce the model’s sensitivity to minor fluctuations. A flattening operation subsequently converts the pooled features into a one-dimensional vector suitable for the final classification stage.
In this adaptation, the original regression objective—predicting BGC values—is replaced by a classification task that identifies the patient corresponding to each input sequence. This modification enables a detailed analysis of how key features—such as CGM, insulin administration, and carbohydrate intake—contribute to individual patient characterization. The architectural refinements enhance both model interpretability and subject differentiation, allowing for more effective capture of unique physiological and behavioral patterns.

\subsubsection*{Loss Function for Patient Identification}
To train the model for patient classification, we adopt a configuration aligned with the original task, using cross-entropy loss. This setup ensures effective learning of discriminative, patient-specific patterns by optimizing the model to distinguish individuals based on their unique physiological and behavioral profiles.

\subsection{Patient-Independent vs Patient-Specific Models} 
As illustrated in Figure~\ref{fig:regression}, we introduce a two-phase training strategy based on a Bi-GRU architecture, designed to address both patient-independent and patient-specific blood glucose prediction tasks. This approach enables the model to capture general temporal dynamics across the population while supporting subject-level adaptation for improved personalization.
%
%

The network processes multimodal input data—including CGM measurements, basal and bolus insulin doses, and carbohydrate intake—through bidirectional recurrent layers, followed by fully connected layers optimized for regression. This configuration allows the model to learn complex temporal relationships and produce accurate BGC forecasts.
In the first phase, the model is trained on a diverse set of patient data to capture generalized glucose dynamics. In the second phase, it is fine-tuned on limited subject-specific data, allowing the model to specialize in individual physiological patterns. This strategy balances generalization and personalization, making it suitable for real-world deployment in personalized diabetes management systems.

\noindent The training process consists of two main phases:

\begin{enumerate}
    \item {Patient-Independent Training}---A generalized Bi-GRU model is trained using LOSOCV, where each subject is iteratively excluded from the training set and used for testing. This systematic rotation ensures robustness and mitigates bias by validating the model under consistent and comprehensive conditions.
    \item {Patient-Specific Fine-Tuning}---The generalized model is adapted to each test subject by further training the model with data from the test subject. We enable it to adjust to individual physiological patterns, enhancing personalization and improving subject-specific performance.
\end{enumerate}

\subsubsection*{Loss Function for Glucose Prediction}
To ensure fair and unbiased glucose level predictions, particularly for rare adverse events, we integrate the shrinkage loss function~\cite{shrinkage} into the training process. This approach aligns with recent methodologies, including that in~\cite{bg-bert} and~\cite{rigamonti2024improving}. Due to the limited occurrence of adverse events, standard loss functions may fail to adequately capture their significance. The shrinkage loss function mitigates this issue by steering the model's focus toward these critical cases. Its formulation is given by:
\begin{equation}
 L_s = \frac{\lVert \hat{g} - g \rVert^2}{1 + \exp(a \cdot (c - \lVert \hat{g} - g \rVert))}
 \end{equation}
where $\hat{g}$ denotes predicted glucose levels, $g$ represents reference glucose levels, and $a$ and $c$ are modulation parameters that adjust the behavior of the shrinkage loss function.


\section{Experiments}
In this section, we outline the experimental setup and the evaluation metrics used to assess our experiments.

\subsection{Materials}

\begin{table*}[b]
\centering
\captionsetup{font=small}
\caption{Summary of total samples, hypo/hyper-glycemic events for each subject in (a) OhioT1DM and (b) DiaTrend datasets.}
\label{tab:summary_datasets}
\begin{subtable}[t]{0.45\textwidth}
\centering
\captionsetup{font=small}
\caption{OhioT1DM Dataset}
\label{tab:ohio} 
\vspace{0.5cm}
\resizebox{\textwidth}{!}{
\begin{tabular}{lccc}
\toprule
\textbf{Sub. ID} & \textbf{Total} & \textbf{Hypo (\%)} & \textbf{Hyper (\%)} \\
\midrule
540 & 402,510 & 6.22\% & 4.93\% \\
544 & 368,130 & 1.41\% & 9.05\% \\
552 & 305,940 & 3.30\% & 4.80\% \\
559 & 338,520 & 3.82\% & 13.45\% \\
563 & 350,760 & 1.92\% & 2.02\% \\
567 & 334,710 & 7.10\% & 6.80\% \\
570 & 359,790 & 1.66\% & 20.51\% \\
575 & 356,910 & 7.66\% & 4.14\% \\
584 & 385,380 & 0.69\% & 15.11\% \\
588 & 391,320 & 0.71\% & 4.86\% \\
591 & 326,670 & 3.16\% & 6.93\% \\
596 & 361,650 & 2.05\% & 3.74\% \\
\midrule
\textit{Total} & 4,282,290 & 3.27\% & 8.06\% \\
\end{tabular}}
\end{subtable}
\hspace{0.05\textwidth}
\begin{subtable}[t]{0.45\textwidth}
\centering
\captionsetup{font=small}
\caption{DiaTrend Dataset}
\label{tab:diatrend} 
\resizebox{\textwidth}{!}{
\begin{tabular}{lccc}
\toprule
\textbf{Sub. ID} & \textbf{Total} & \textbf{Hypo (\%)} & \textbf{Hyper (\%)} \\
\midrule
29 & 1,366,020 & 3.20\% & 8.72\% \\
30 & 1,358,580 & 1.60\% & 15.44\% \\
31 & 1,471,770 & 0.43\% & 9.08\% \\
36 & 767,460  & 0.21\% & 16.53\% \\
37 & 813,150 & 0.52\% & 12.60\% \\
38 & 765,090 & 0.72\% & 4.12\% \\
39 & 728,070 & 1.51\% & 2.26\% \\
42 & 638,880 & 1.69\% & 3.80\% \\
45 & 406,680 & 0.95\% & 2.61\% \\
46 & 327,720 & 2.30\% & 22.90\% \\
47 & 314,190 & 0.07\% & 0.44\% \\
49 & 259,800 & 2.56\% & 7.29\% \\
50 & 230,490 & 0.68\% & 35.64\% \\
51 & 244,890 & 1.25\% & 17.27\% \\
52 & 233,700 & n.a.    & 21.26\% \\
53 & 259,980 & 3.16\% & 12.06\% \\
54 & 225,000 & 3.31\% & 0.43\% \\
\midrule
\textit{Total} & 10,411,470 & 1.38\% & 10.34\% \\
\end{tabular}}
\end{subtable}
\end{table*}

We conduct all experiments on two widely used benchmark datasets, OhioT1DM~\cite{ohio} and DiaTrend~\cite{diatrend}, both offering rich multimodal data combining CGM with contextual information.
OhioT1DM includes data from 12 individuals—six from the 2018 challenge and six from 2020—with CGM sampled every five minutes over eight weeks, along with optional fingerstick values, insulin records, physiological signals, and self-reported lifestyle factors.
DiaTrend contains CGM and insulin pump data from 54 individuals with T1D, totaling over 27,000 days of CGM and 8,000 days of pump data, with detailed logs of insulin, carbohydrate intake, pump settings, and demographic and clinical information. We focus on 17 participants with complete basal insulin profiles.

\subsection{Data Balancing and Splitting}

Since hypo- and hyperglycemic events are relatively infrequent in both datasets—3.27\% and 8.06\% in the OhioT1DM dataset, and 1.38\% and 10.34\% in the DiaTrend dataset, respectively—a stratified data split is used to maintain a balanced distribution of event types. Specifically, each subject’s data is divided into training (64\%), validation (16\%), and test (20\%) sets while preserving event proportions. These are aggregated across subjects to form well-balanced overall datasets, supporting robust model training and evaluation. Tables~\ref{tab:ohio} and~\ref{tab:diatrend} provide detailed sample and event distributions per subject.


\subsection{Preprocessing}
To ensure the integrity and suitability of our datasets, we implement a comprehensive preprocessing pipeline. First, we synchronize the features with BGC readings following the approach in~\cite{bg-bert} and~\cite{rigamonti2024improving}, aligning each glucose timestamp with subsequent observations within a four-minute window to minimize temporal misalignment.
We then construct individual time series representing continuous glucose monitoring intervals. We apply linear interpolation to maintain a uniform five-minute interval between observations and to fill minor gaps due to missing glucose readings. However, if a gap exceeded 20 minutes (i.e., five missing samples), we split the sequence instead of imputing data, generating a new time series.
We ensure continuity in representing bolus insulin and carbohydrate intake following the methodology in~\cite{butt2023feature}. Bolus insulin data is processed using an exponential decay curve to estimate insulin on board, while carbohydrate intake is modeled with a customized function to reflect absorption rates over time, capturing dynamic changes post-ingestion.
Finally, we apply min-max normalization to scale all features uniformly, ensuring consistency across the time series.

\subsection{Experimental Configuration}
All experiments in this study use the two datasets described in the previous section. 
For the patient identification task, the input consists of a window of 24 consecutive timestamps (120 mins) of observed data. Each input window includes multiple physiological and behavioral features (i.e., CGM readings, basal insulin, bolus insulin, and carbohydrate intake). This multimodal representation provides a comprehensive snapshot of the patient’s glycemic and insulin response patterns over time, allowing the classification model to effectively distinguish between individuals based on their unique physiological characteristics.
For the BGC prediction task, we employ two different configurations for each dataset, varying the observation and prediction windows. The first configuration consists of a 30-timestamp window (150 mins), with 24 timestamps (120 mins) for observed data and six (30 mins) for predictions. The second configuration extends to a 60-timestamp window (300 mins), with 48 timestamps (240 mins) of observed data and 12 timestamps (60 mins) for forecasts. These setups are referred to as OhioT1DM (PH = 30 mins) and DiaTrend (PH = 30 mins) for the first configuration; OhioT1DM (PH = 60 mins) and DiaTrend (PH = 60 mins) for the second one.
Since all configurations exhibit a significant imbalance between normal and adverse BGCs, we use the Synthetic Minority Oversampling Technique (SMOTE)~\cite{chawla2002smote} exclusively to the regression task to balance the training data. This approach enhances the representation of adverse BGCs by generating synthetic samples based on k-nearest neighbors, thereby improving BGC prediction.

\subsection{Comparative Analysis and Ablation Study}
To contextualize our results, we compare our glucose prediction models against the state-of-the-art baseline proposed by~\cite{deng2021deep}, which utilizes TimeGAN for data augmentation and a CNN-based pretraining and fine-tuning approach for personalization. Since this baseline relies exclusively on CGM data, we ensure a fair comparison by evaluating our Bi-GRU model under the same conditions.
Unlike the baseline, which directly predicts glucose values at a specific prediction horizon, our approach estimates changes in BGC over varying Prediction Horizons (PH). Particularly, we predict the difference in BGC between time points $t$ and $t+PH$. To maintain fairness, we adapt the baseline models to follow this formulation, enabling a direct and equitable comparison. To the best of our knowledge, this baseline represents the current state-of-the-art in personalized glucose prediction models.

Building on our patient identification results, which demonstrate that CGM readings, basal insulin, bolus insulin, and carbohydrate intake effectively distinguish individuals based on their unique physiological profiles, we extend our evaluation beyond CGM-only benchmarks. Specifically, we assess our forecasting approach within a multimodal framework to determine whether integrating other data sources enhances predictive accuracy and robustness. This evaluation provides deeper insights into the role of multimodal fusion in both predictive performance and personalization in BGC forecasting.

Additionally, we conduct an ablation study to examine how training with limited patient-specific data affects accuracy. This study uses the multimodal approach, highlighting the importance of incorporating all features in patient-specific models. The analysis offers valuable insights into the model’s ability to learn from smaller datasets, helping to identify the minimum data required for reliable glucose prediction, as further explored in the Results section. This aspect is particularly critical for real-world applications where extensive data collection may not always be feasible.


\subsection{Training Configuration}
For training all experiments, we maintained the original hyperparameter values from~\cite{bg-bert}, using a batch size of 1024 and an initial learning rate of $2\mathrm{e}{-4}$. Each experiment, both in the benchmark and ablation studies, was conducted over 3200 epochs with early stopping applied at a patience level of 200. We employed the Adam optimizer with a learning rate of $1e^{-4}$ and used a Reduce Learning Rate on Plateau scheduler with a patience of 15 and a reduction factor of 0.5.




\subsection{Analytical performance assessment criteria} 
\subsubsection*{Patient Identification}
To evaluate the effectiveness of our approach in the patient identification task, we employ standard classification metrics, including Accuracy, Precision, Recall, and F1-score. These metrics provide a comprehensive assessment of the model's ability to correctly distinguish between individual patients based on their glucose dynamics.
\subsubsection*{Glucose Prediction}
The analytical evaluation of glucose prediction experiments is based on four widely used metrics in BGC forecasting~\cite{bg-bert,rigamonti2024improving}: Root Mean Square Error (RMSE), Sensitivity on Hyperglycemic Events (Hyper Sen), Sensitivity on Hypoglycemic Events (Hypo Sen), and Time Gain (TG).
The RMSE (Equation~\ref{eq:rmse}) quantifies the root mean square difference between predicted glucose levels $\hat{g}_i$ and reference glucose levels $g_i$: 
\begin{equation} \operatorname{RMSE}(g, \hat{g}) = \sqrt{\frac{1}{n}\sum_{i=1}^{n}(g_i - \hat{g}_i)^2} \label{eq:rmse} 
\end{equation}

Meanwhile, Hyper Sen and Hypo Sen (Equations~\ref{eq:sen_hyper} and \ref{eq:sen_hypo}) measure the model’s ability to accurately detect hyperglycemic and hypoglycemic events, respectively: 

\begin{equation} \operatorname{Hyper\ Sen} = \frac{\mathit{True Hyper Events}}{\mathit{True Hyper Events} + \mathit{Missed Hyper Events}} \label{eq:sen_hyper} 
\end{equation}

\begin{equation} \operatorname{Hypo\ Sen} = \frac{\mathit{True Hypo Events}}{\mathit{True Hypo Events} + \mathit{Missed Hypo Events}} \label{eq:sen_hypo} 
\end{equation}

\noindent where $\mathit{True Hyper Events}$ and $\mathit{True Hypo Events}$ represent the hyperglycemic and hypoglycemic events correctly identified by the model, while $\mathit{Missed Hyper Events}$ and $\mathit{Missed Hypo Events}$ denote events that the model failed to detect.

Lastly, TG (Equation~\ref{eq:tg}) measures the average time gained for early detection of potential adverse events using the model: 
\begin{equation} \operatorname{TG}(g,\hat{g}) = \mathit{PH} - {\operatorname{delay}}(g,\hat{g}) \label{eq:tg} \end{equation}

\noindent where $\mathit{PH}$ is the prediction horizon, and delay represents the time shift $k$ that minimizes the distance between the original and predicted glucose profiles: 
\begin{equation} {\operatorname{delay}}(g, \hat{g}) = \underset{k}{\mathrm{argmin}}\sum_{i=1}^L (g_{i} - \hat{g}_{i - k})^2 \end{equation}


\begin{figure}[ht] 
    \centering
        \begin{subfigure}[t]{0.52\linewidth}
            \centering
            \includegraphics[width=\linewidth]{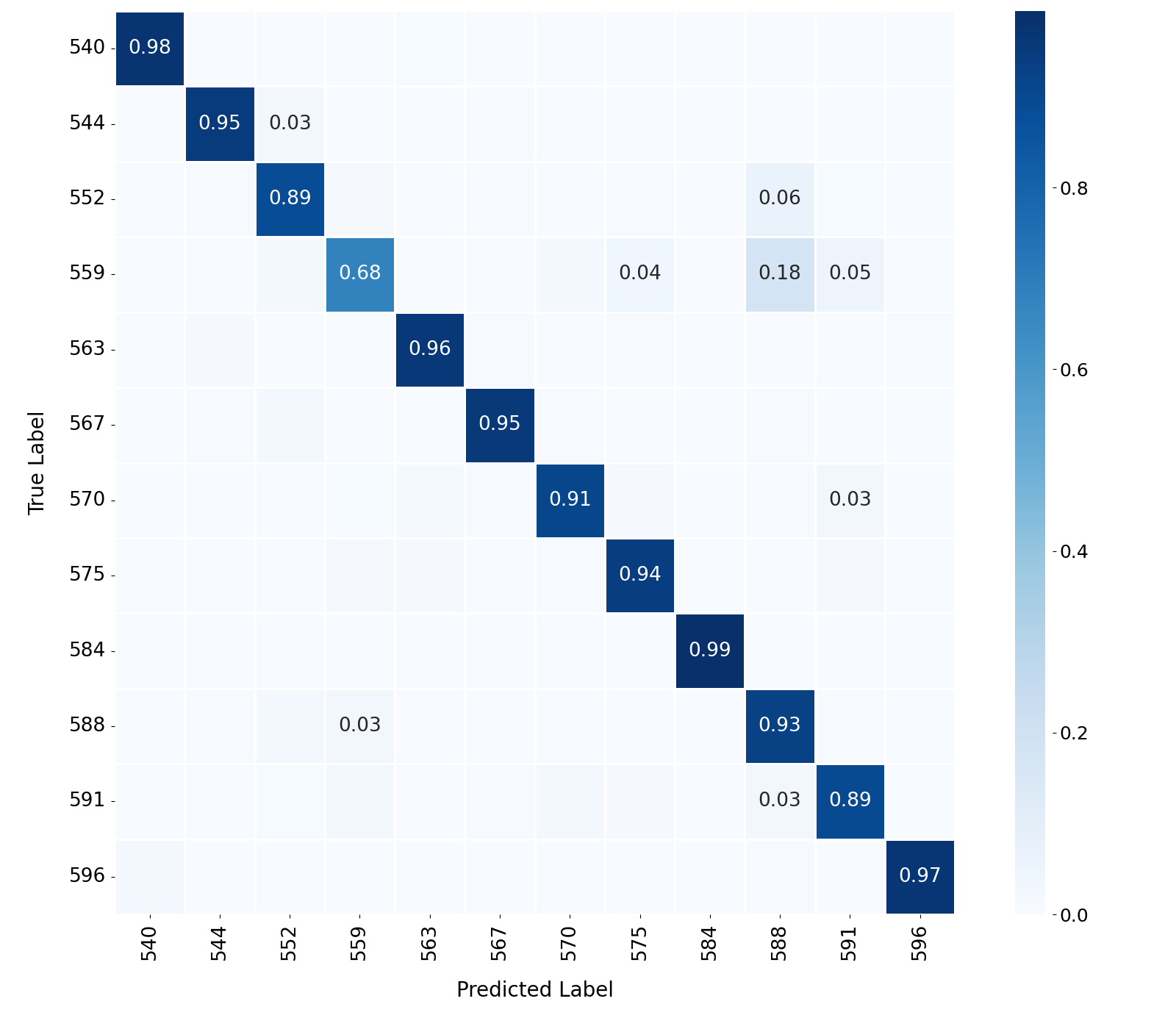}
            \subcaption{OhioT1DM dataset} \label{fig:ohio_confusion}
        \end{subfigure}
        \vspace{1em}
        \begin{subfigure}[t]{0.57\linewidth}
            \centering
            \includegraphics[width=\linewidth]{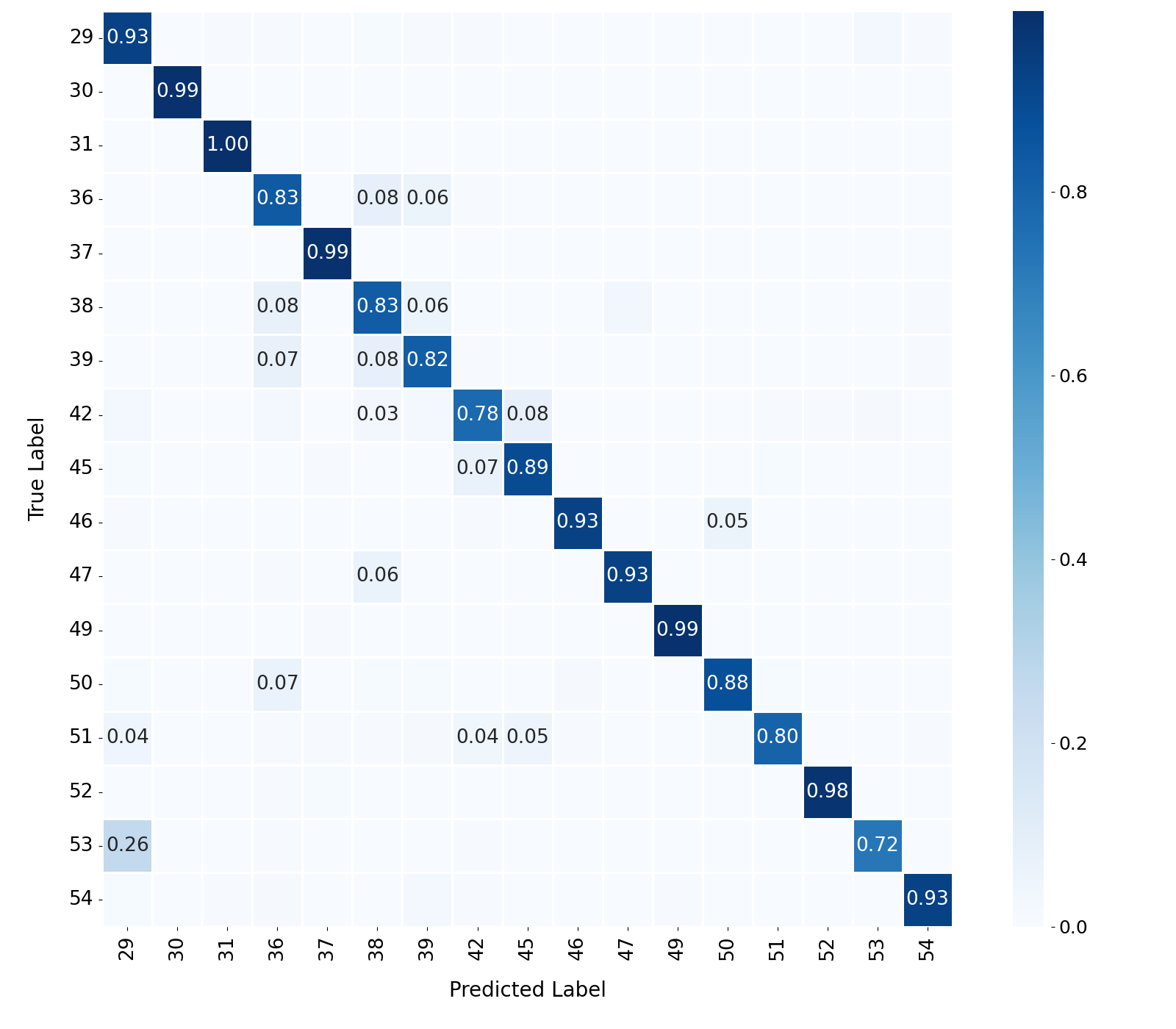}
            \subcaption{DiaTrend dataset} \label{fig:diatrend_confusion}
        \end{subfigure}
    
    \caption{Confusion matrices for the patient identification task using (a) the OhioT1DM and (b) the DiaTrend datasets. Darker shades indicate higher classification accuracy.}
    \label{fig:confusion_matrices}
\end{figure}


\section{Results and Discussions}
This section presents the results of patient identification and BGC prediction, focusing on model performance.

\subsection{Patient Identification Results}
Figures~\ref{fig:ohio_confusion} and~\ref{fig:diatrend_confusion} show the confusion matrices for the OhioT1DM and DiaTrend datasets respectively, providing insight into the performance of the patient recognition classifier. 
Both confusion matrices show a strong diagonal pattern, indicating that the classifier accurately identifies patients in most cases. Although some misclassifications occur due to difficulties distinguishing certain patterns, the overall accuracy remains high, reaching about 92\% for the OhioT1DM dataset and 91\% for the DiaTrend dataset.
In the OhioT1DM dataset, the classifier achieves high accuracy for patients 540, 584, and 596, with true positive rates of 98\%, 99\%, and 97\%, respectively. In contrast, performance drops for others, such as patient 559, with a true positive rate of 68\%, indicating increased misclassification.
Similarly, in the DiaTrend dataset, strong results are observed for patients 30, 31, 37, 49, and 52, with true positive rates ranging from 98\% to 100\%.  However, lower accuracy is noted for patients 42 (78\%), 51 (80\%), and 53 (72\%), reflecting variability in classification performance across individuals.
These errors suggest that certain patients are more challenging to distinguish, possibly due to similar characteristics or overlapping representations in the model’s latent space.
Despite minor misclassifications, the classifier exhibits performance, with precision, recall, and F1-scores of 92\% for OhioT1DM and 90\%, 89\%, and 90\% for DiaTrend.

\subsection{Patient-Independent vs. Patient-Specific Models Results}
Table~\ref{table:results} presents the performance across subjects for patient-independent and patient-specific methods on OhioT1DM and DiaTrend, evaluated under two PHs (30 and 60 mins). This comparison examines the efficacy of our proposed Bi-GRU model and training strategy against the baseline CNN model, which use an unimodal approach, relying on CGM data only. We also assess the impact of incorporating physiological features (basal insulin, bolus insulin, and carbohydrate intake) alongside CGM data, transitioning to a multimodal framework to improve accuracy.
The results reveal an advantage of patient-specific models over patient-independent models across datasets and PHs. This consistent trend highlights the importance of personalized modeling, as tailoring the model to an individual’s glucose dynamics significantly enhances prediction accuracy and improves sensitivity to extreme glucose fluctuations.
Beyond personalization, a key finding is the superior performance of Bi-GRU over CNN across all configurations within the unimodal setting. The advantage of Bi-GRU is particularly evident in RMSE, TG, and Hypo Sen, indicating that recurrent models are better suited to capture temporal dependencies in BGC data. Unlike CNNs, which primarily extract spatial features, Bi-GRUs leverage sequential processing to capture long-term glucose fluctuations more effectively, yielding more accurate and stable predictions.
Although these findings demonstrate the effectiveness of patient-specific modeling and recurrent networks within a CGM-only framework, a greater performance boost is observed when transitioning to a multimodal approach. Incorporating additional physiological features consistently enhances RMSE, TG, and Hypo Sen, particularly at the PH of 60 mins, highlighting the importance of context-aware glucose prediction models in improving both accuracy and robustness. By providing contextual information about external factors influencing glucose variations, multimodal models enhance the accuracy and timeliness of glucose trend predictions, demonstrating their potential clinical utility in diabetes management.

\begin{table*}[tb]
\centering
\caption{Patient-Independent vs. Patient-Specific models Results\textsuperscript{*}}
\resizebox{1.0\textwidth}{!}{
\begin{tabular}{llccccccccccc}
    \toprule
    & & & \multicolumn{5}{c}{PH = 30 mins} & \multicolumn{4}{c}{PH = 60 mins}\\  
    \cmidrule(lr){5-8} \cmidrule(lr){9-12}
    Dataset & Model & Patient-Specific & Features & RMSE & TG & Hyper Sen & Hypo Sen & RMSE & TG & Hyper Sen & Hypo Sen\\
    & & & & (mg/dL)↓ & (mins)↑ & (\%)↑  & (\%)↑ & (mg/dL)↓ & (mins)↑ & (\%)↑  & (\%)↑\\
    \midrule
\multirow{6}{*}{OhioT1DM} & CNN & - & - & 14.34 & 15.74 & \textbf{84.43} & 73.41 & 23.81 & 27.99 & 65.84 & 47.34 \\ 
& CNN & \cmark & - & 14.18 & 15.92 & 83.03 & 70.26 & 23.32 & 28.61 & 65.94 & 45.36 \\
& Bi-GRU & - & - & 14.14 & 17.24 & 82.22 & 74.01 & 25.20 & 31.88 & 63.69 & 42.39 \\ 
& Bi-GRU (our) & \cmark & - & 14.86 & 18.41 & 83.43 & 75.52 & 22.27 & 34.00 & 66.05 & 59.30 \\
& Bi-GRU & - & \cmark & \textbf{14.05} & 17.79 & 82.06 & 70.73 & 27.99 & 33.96 & 56.03 & 29.47 \\
& Bi-GRU (our) & \cmark & \cmark & 14.15 & \textbf{18.97} & 84.10 & \textbf{76.01} & \textbf{19.39} & \textbf{36.71} & \textbf{68.03} & \textbf{71.61} \\
\midrule
\multirow{6}{*}{DiaTrend} & CNN & - & - & 15.35 & 15.35 & \textbf{83.02} & 51.30 & 24.91 & 29.12 & \textbf{68.98} & 26.70\\
& CNN & \cmark & - & 15.05 & 15.65 & 79.75 & 49.68 & 24.34 & 28.56 & 66.48 & 28.25 \\    
& Bi-GRU & - & - & 15.24 & 16.55 & 79.12 & 52.96 & 27.28 & 33.01 & 61.81 & 21.26 \\
& Bi-GRU (our) & \cmark & - & 15.63 & 17.90 & 81.29 & 57.69 & 23.77 & 34.98 & 67.47 & 43.82 \\
& Bi-GRU & - & \cmark & 15.19 & 16.93 & 79.39 & 48.70 & 30.89 & 34.47 & 56.45 & 9.27 \\
& Bi-GRU (our) & \cmark & \cmark & \textbf{14.88} & \textbf{17.95} & 81.15 & \textbf{60.33} & \textbf{21.59} & \textbf{35.70} & 66.77 & \textbf{49.59} \\
\bottomrule
    \end{tabular}}
    \\
    \vspace{0.5em}
    \raggedright{
    \hspace{0.9em}\scriptsize{\textsuperscript{*} The values shown are average results. In \textbf{bold} the best score.}}\\
\label{table:results}
\end{table*}

\subsection{Ablation Study}

\begin{figure*}[h]
    \centering
    \captionsetup[subfigure]{
        font=footnotesize,
        justification=justified, 
        singlelinecheck=off,
        format=hang}

    \begin{minipage}{0.48\textwidth}
        \centering
        \begin{subfigure}[t]{0.48\textwidth}
            \centering
            \includegraphics[width=\linewidth]{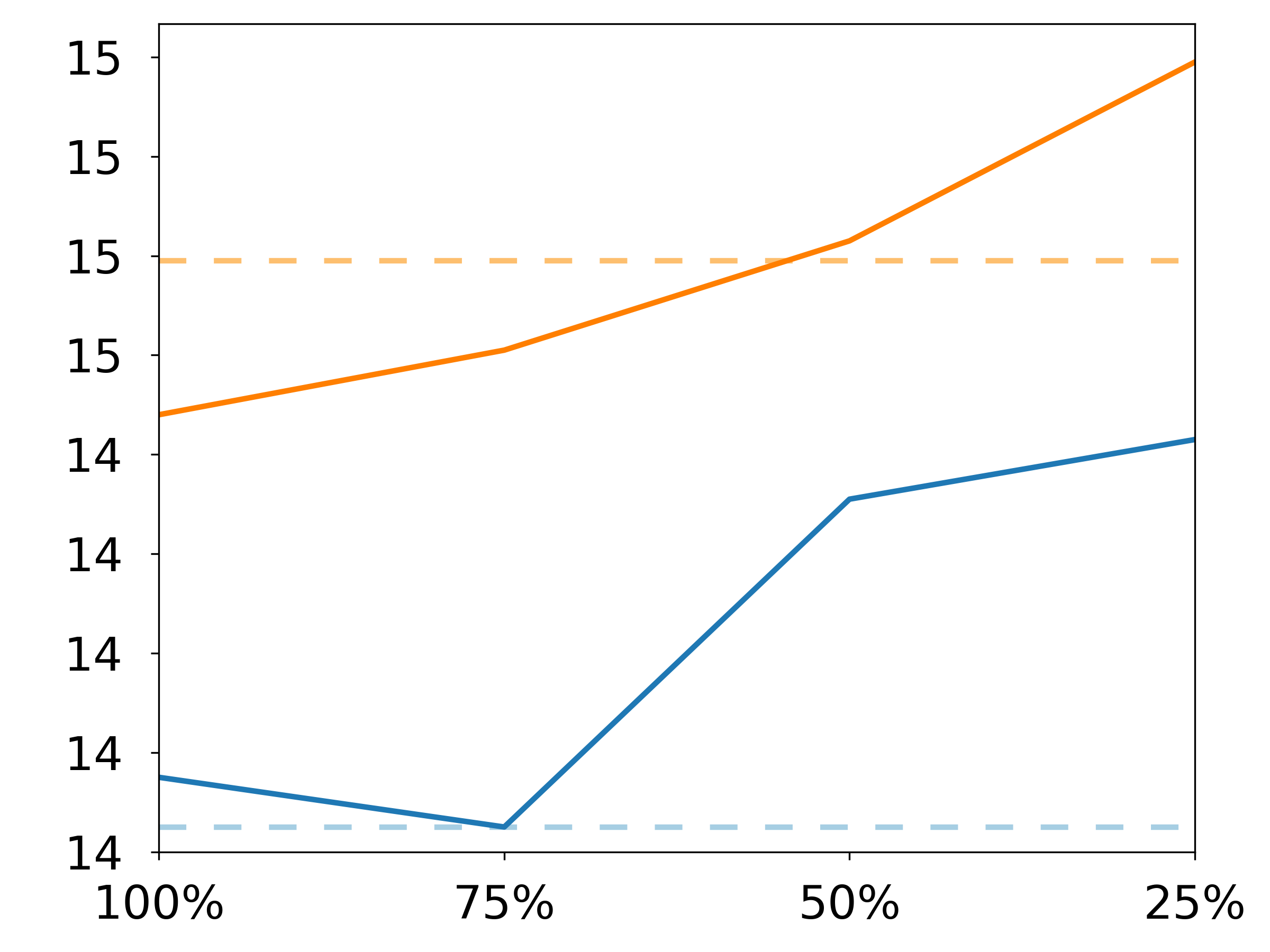}
            \caption[a]{RMSE\\(PH=30 mins)}
        \end{subfigure}
        \hfill
        \begin{subfigure}[t]{0.48\textwidth}
            \centering
            \includegraphics[width=\linewidth]{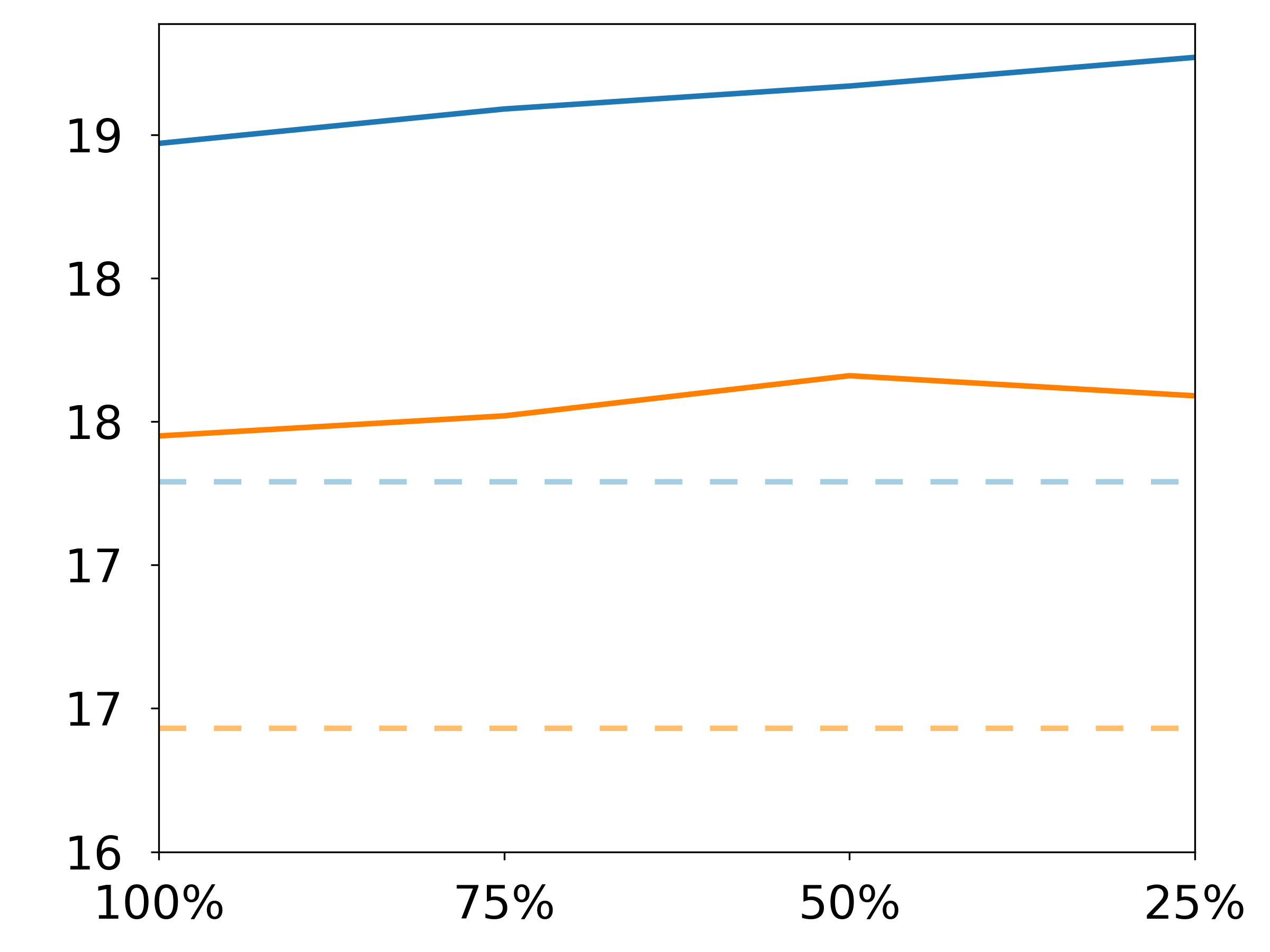}
            \caption[b]{TG\\(PH=30 mins)}
        \end{subfigure}

        \vspace{1em}

        \begin{subfigure}[t]{0.48\textwidth}
            \centering
            \includegraphics[width=\linewidth]{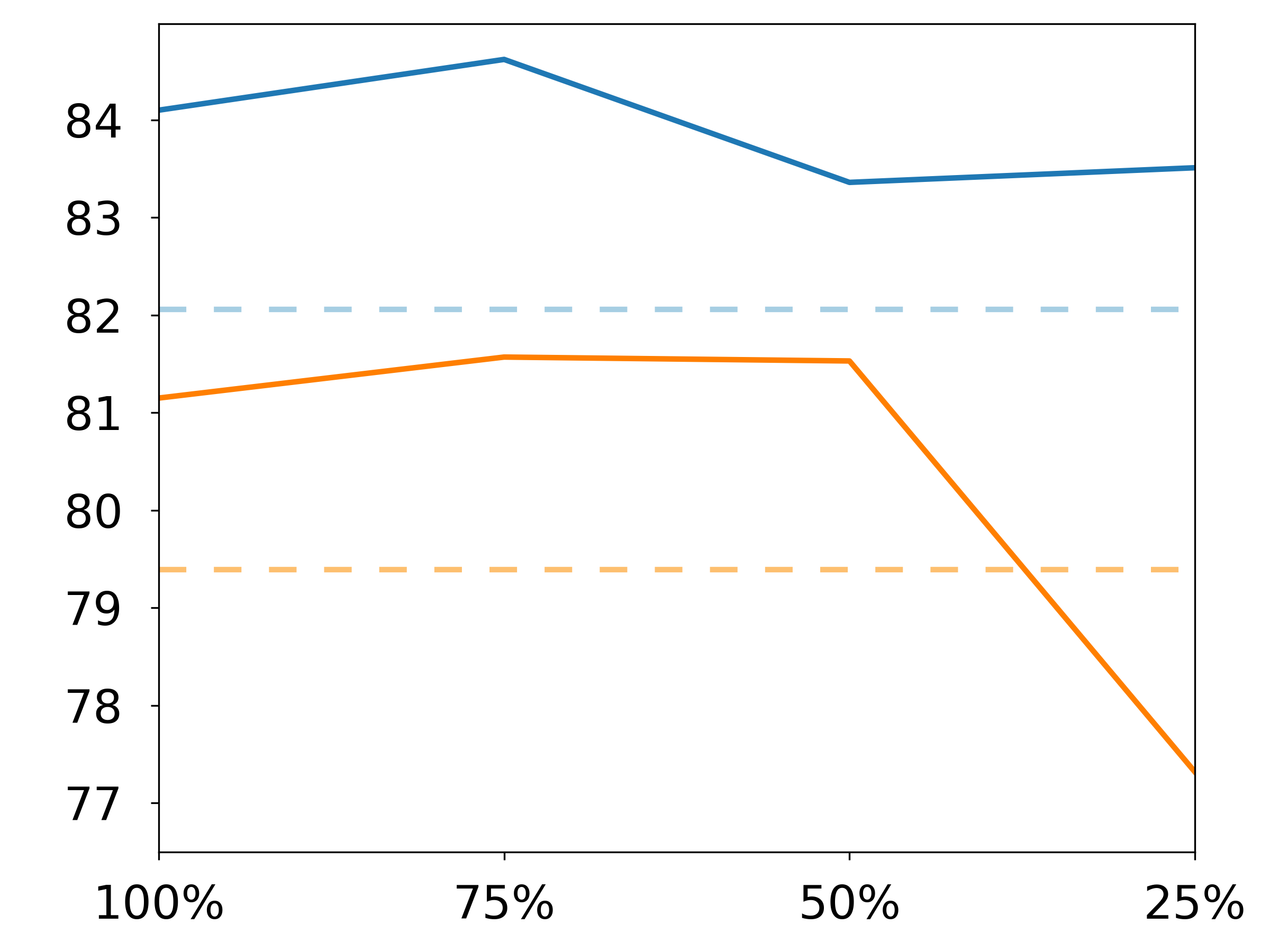}
            \caption[c]{Hyper Sen\\(PH=30 mins)}
        \end{subfigure}
        \hfill
        \begin{subfigure}[t]{0.48\textwidth}
            \centering
            \includegraphics[width=\linewidth]{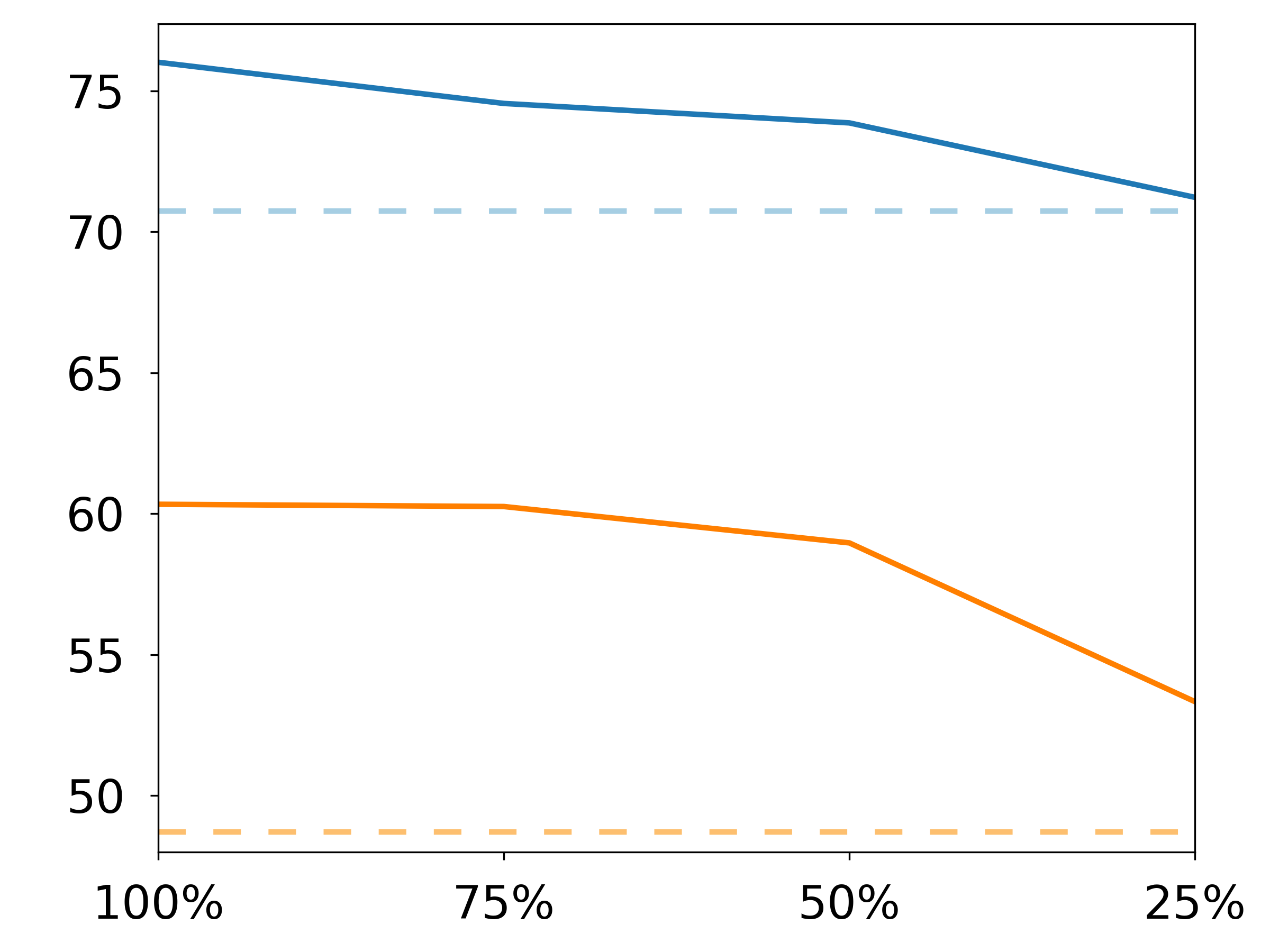}
            \caption[d]{Hypo Sen\\(PH=30 mins)}
        \end{subfigure}
    \end{minipage}
    \hfill
    \begin{minipage}{0.48\textwidth}
        \centering
        \begin{subfigure}[t]{0.48\textwidth}
            \centering
            \includegraphics[width=\linewidth]{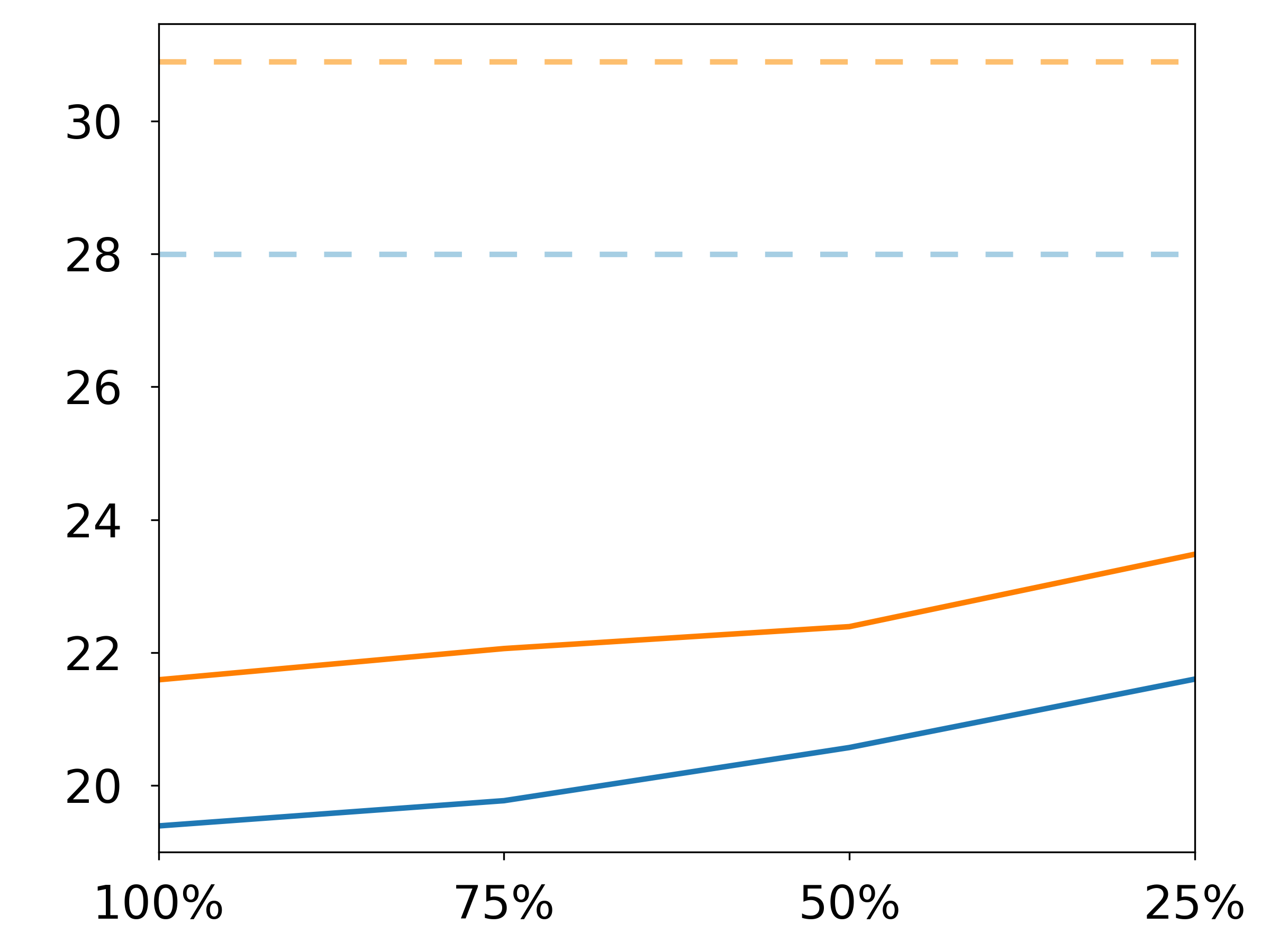}
            \caption[e]{RMSE\\(PH=60 mins)}
        \end{subfigure}
        \hfill
        \begin{subfigure}[t]{0.48\textwidth}
            \centering
            \includegraphics[width=\linewidth]{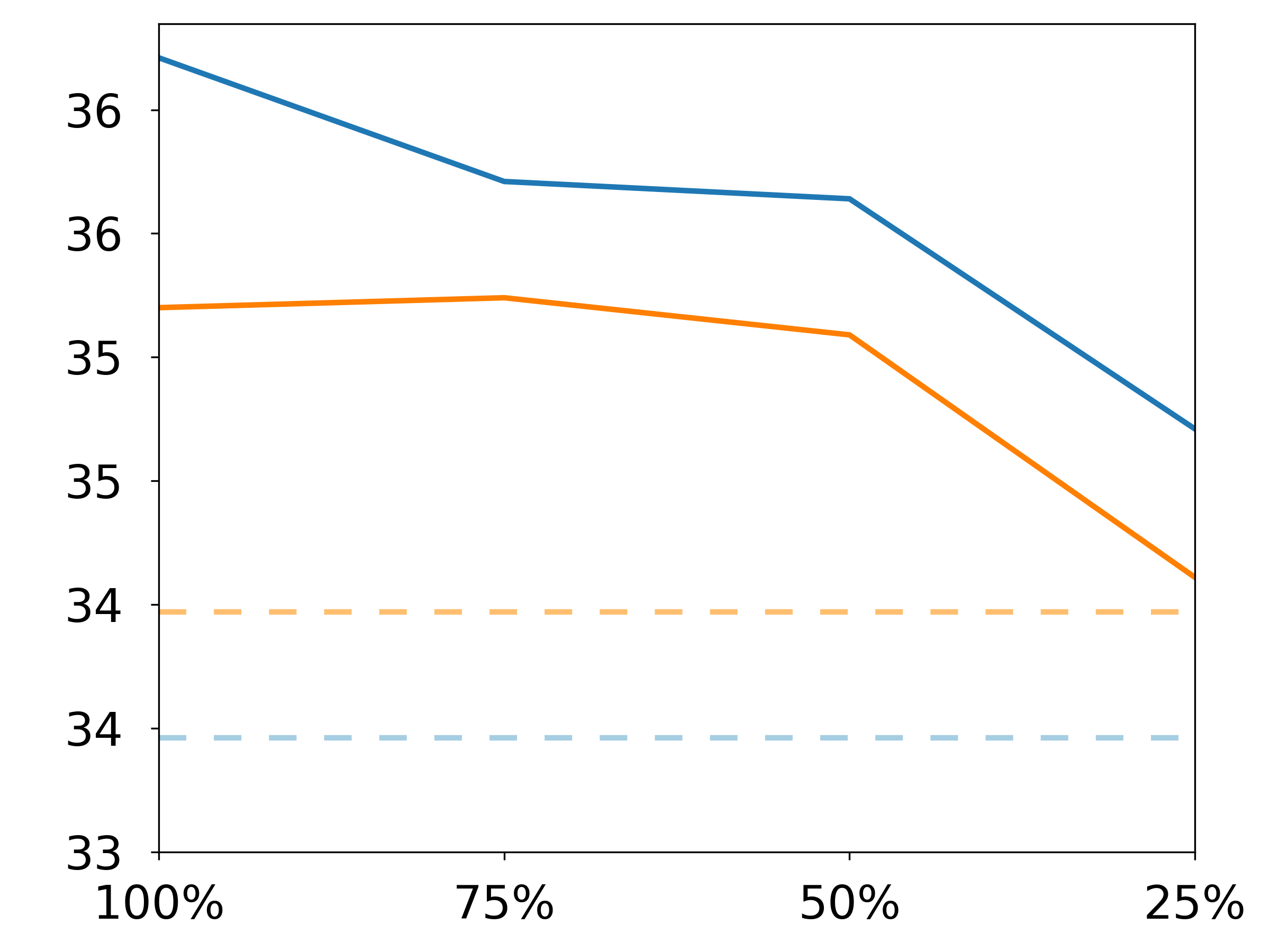}
            \caption[f]{TG\\(PH=60 mins)}
        \end{subfigure}

        \vspace{1em}

        \begin{subfigure}[t]{0.48\textwidth}
            \centering
            \includegraphics[width=\linewidth]{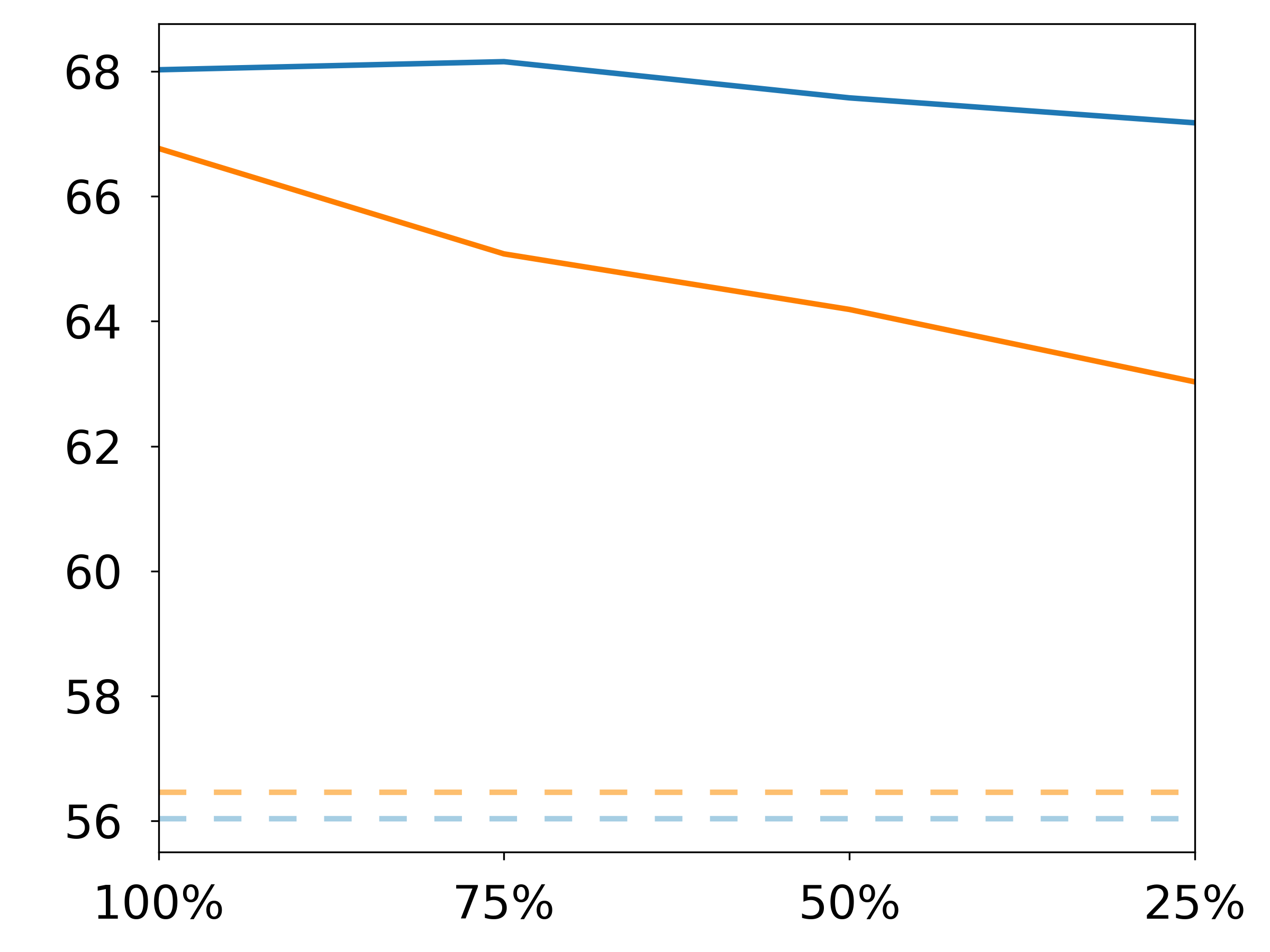}
            \caption[g]{Hyper Sen\\(PH=60 mins)}
        \end{subfigure}
        \hfill
        \begin{subfigure}[t]{0.48\textwidth}
            \centering
            \includegraphics[width=\linewidth]{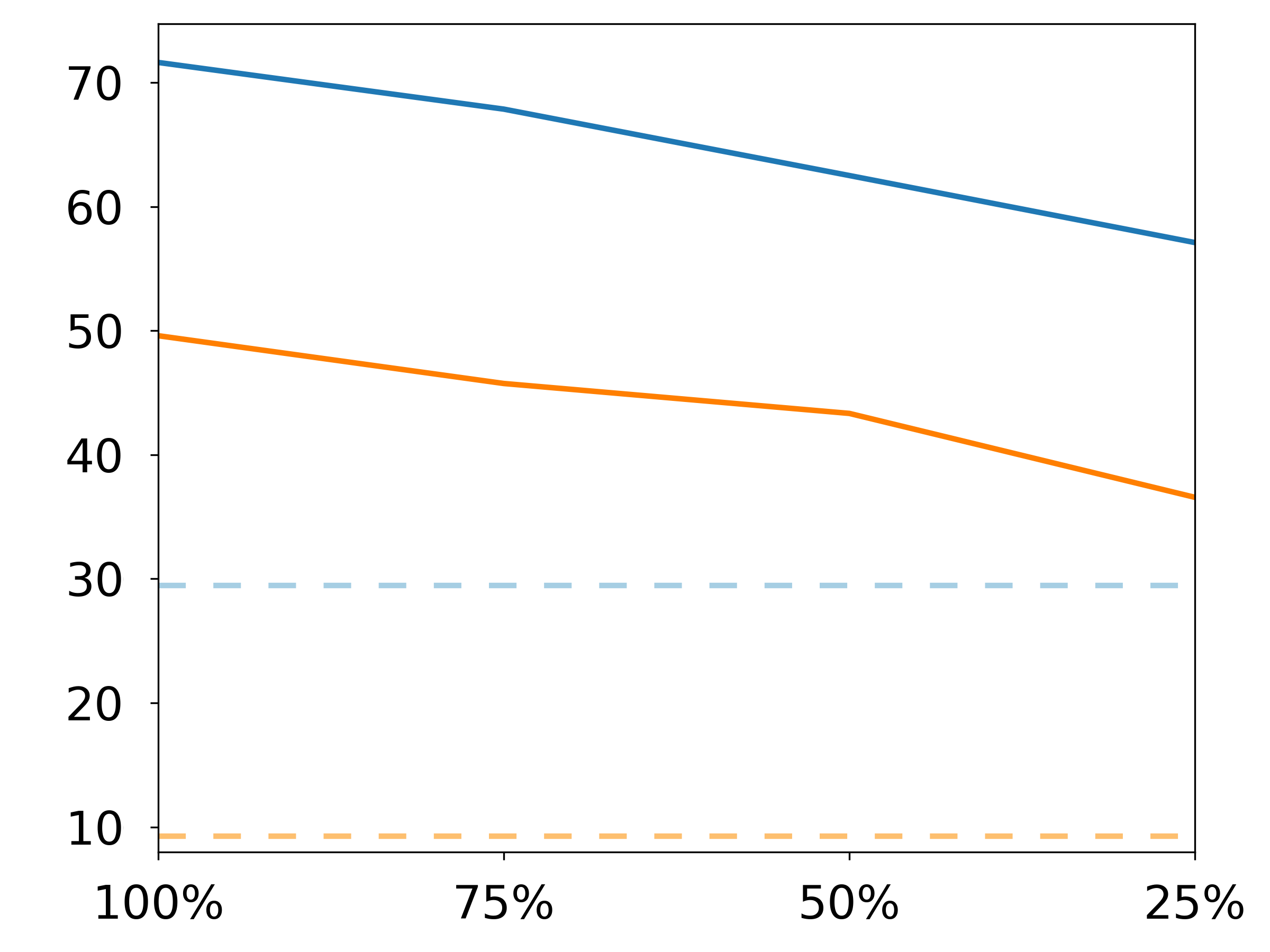}
            \caption[h]{Hypo Sen\\(PH=60 mins)}
        \end{subfigure}
    \end{minipage}

    \caption{\small Ablation study on patient-specific models with decreasing training data. (a--d) 30-min prediction horizon; (e--h) 60-min prediction horizon. Light orange and light blue dashed lines represent patient-independent models trained on OhioT1DM and DiaTrend, respectively. Solid orange and blue lines correspond to patient-specific models trained on the same datasets.}
    \label{fig:ablation_study}
\end{figure*}

Training a patient-independent model from scratch requires significantly more time than adapting a patient-specific model. 
In OhioT1DM, training a patient-independent model takes, on average, six times longer than personalizing a patient-specific model, with an average personalization data span of 27 days—though fewer days may suffice. For instance, patient 552 achieved highly positive results with just 23 days of data.
Similarly, in DiaTrend, training a patient-independent model takes, on average, four times longer than adapting a patient-specific model. The typical personalization period spans 45 days, yet patient 54 demonstrated strong results with as little as 17 days of data.
To evaluate the impact of patient-specific data availability on model performance, we conducted an ablation study by gradually reducing the amount of train data. Models are trained using 75\%, 50\%, and 25\% of each patient's randomly extracted data (including CGM function, bolus insulin, basal insulin, and carbohydrate intake). This study is performed on both datasets, with performance assessed for PH of 30 and 60 minutes. This approach allows for an analysis of model adaptability across different forecasting windows. Figure~\ref{fig:ablation_study} presents the average results across all subjects.
This analysis highlights the model’s ability to learn from gradually smaller datasets, helping to identify the minimum amount of personalized data needed for effective glucose prediction—a crucial factor given real-world data collection challenges. While patient-specific models perform best with full datasets, they remain effective even with reduced data, though performance depends on both the data and personal characteristics. As training data decreases, performance declines, though the extent varies across patients. Some require larger datasets, while others maintain accuracy with less data, reflecting individual differences in data efficiency. 

\section{Conclusion}

In this study, we investigated how daily system log data influence individual glucose trends in people with T1D. We found that variability in these logs is highly distinctive and can reliably identify individual patients, emphasizing the importance of incorporating patient-specific patterns into BGC prediction and insulin management. This observation supports the broader argument that personalization is essential for accurate forecasting in real-world scenarios.
Our results show that adapting a pre-trained deep learning model to new patients is significantly more practical and efficient than training from scratch. This approach enables continuous adaptation, reduces computational demands and training time, and maintains high accuracy, particularly in detecting adverse glucose events such as hypoglycemia and hyperglycemia. By comparing patient-independent and patient-specific models, we demonstrate that the latter consistently outperforms the former across multiple metrics. This patient-specific strategy aligns with the principles of precision medicine, optimizing therapeutic interventions based on individual physiological and behavioral profiles.
An ablation study confirms that strong predictive performance can be retained even with reduced amounts of subject-specific data—an important consideration when dealing with incomplete or noisy real-world data. Furthermore, integrating multimodal inputs such as insulin delivery, carbohydrate intake, and contextual signals significantly improves forecast accuracy, especially for longer prediction horizons. While these results are promising, challenges remain for patients with highly irregular glucose patterns. Future work will focus on refining personalization techniques, scaling to diverse populations, and incorporating additional physiological and behavioral features to improve adaptability and clinical impact.

\section*{Data Availability}
The benchmark datasets OhioT1DM and DiaTrend used in this research are available upon request from the respective authors. For quick reference OhioT1DM is available at \url{https://webpages.charlotte.edu/rbunescu/data/ohiot1dm/OhioT1DM-dataset.html} and DiaTrend is available at \url{https://doi.org/10.7303/syn38187184}.

\section*{Acknowledgements}
This work was funded by the National Plan for NRRP Complementary Investments (PNC, established with the decree-law 6 May 2021, n. 59, converted by law n. 101 of 2021) in the call for the funding of research initiatives for technologies and innovative trajectories in the health and care sectors (Directorial Decree n. 931 of 06-06-2022) - project n. PNC0000003 - AdvaNced Technologies for Human-centrEd Medicine (project acronym: ANTHEM)\footnote{\url{https://fondazioneanthem.it/}}. This work reflects only the authors’ views and opinions, neither the Ministry for University and Research nor the European Commission can be considered responsible for them.






\bibliography{sn-bibliography}

@article{insulin,
  title={Insulin signalling and the regulation of glucose and lipid metabolism},
  author={Saltiel, Alan R and Kahn, C Ronald},
  journal={Nature},
  volume={414},
  number={6865},
  pages={799--806},
  year={2001},
  publisher={Nature Publishing Group UK London}
}

@incollection{Mathew2023,
  title     = {Blood Glucose Monitoring},
  author    = {Mathew, Thomas K and Zubair, Muhammad and Tadi, Prasanna},
  booktitle = {StatPearls},
  year      = {2023},
  publisher = {StatPearls Publishing},
  address   = {Treasure Island (FL)}
}

@article{oviedo2017review,
  title={A review of personalized blood glucose prediction strategies for T1DM patients},
  author={Oviedo, Silvia and Veh{\'\i}, Josep and Calm, Remei and Armengol, Joaquim},
  journal={International journal for numerical methods in biomedical engineering},
  volume={33},
  number={6},
  pages={e2833},
  year={2017},
  publisher={Wiley Online Library}
}

@article{aliberti2019multi,
  title={A multi-patient data-driven approach to blood glucose prediction},
  author={Aliberti, Alessandro and Pupillo, Irene and Terna, Stefano and Macii, Enrico and Di Cataldo, Santa and Patti, Edoardo and Acquaviva, Andrea},
  journal={Ieee Access},
  volume={7},
  pages={69311--69325},
  year={2019},
  publisher={IEEE}
}

@article{podobnik2025metabolic,
  title={Metabolic interventions as adjunctive therapies to insulin in type 1 diabetes: Current clinical landscape and perspectives},
  author={Podobnik, Juliana and Prentice, Kacey J},
  journal={Diabetes, Obesity and Metabolism},
  year={2025},
  publisher={Wiley Online Library}
}

@article{hyper,
  title={Perioperative hyperglycemia and risk of adverse events among patients with and without diabetes},
  author={Kotagal, Meera and Symons, Rebecca G and Hirsch, Irl B and Umpierrez, Guillermo E and Dellinger, E Patchen and Farrokhi, Ellen T and Flum, David R and others},
  journal={Annals of surgery},
  volume={261},
  number={1},
  pages={97--103},
  year={2015},
  publisher={LWW}
}

@article{hypoglycemia,
  title={Hypoglycemia in diabetes},
  author={Cryer, Philip E and Davis, Stephen N and Shamoon, Harry},
  journal={Diabetes care},
  volume={26},
  number={6},
  pages={1902--1912},
  year={2003},
  publisher={Am Diabetes Assoc}
}

@article{type_1,
  title={Type 1 diabetes},
  author={Atkinson, Mark A and Eisenbarth, George S and Michels, Aaron W},
  journal={The lancet},
  volume={383},
  number={9911},
  pages={69--82},
  year={2014},
  publisher={Elsevier}
}

@article{cgm,
  title={Continuous glucose monitoring: a review of successes, challenges, and opportunities},
  author={Rodbard, David},
  journal={Diabetes technology \& therapeutics},
  volume={18},
  number={S2},
  pages={S2--3},
  year={2016},
  publisher={Mary Ann Liebert, Inc. 140 Huguenot Street, 3rd Floor New Rochelle, NY 10801 USA}
}

@article{csii,
  title={Continuous subcutaneous insulin infusion: a comprehensive review of insulin pump therapy},
  author={Lenhard, M James and Reeves, Grafton D},
  journal={Arch. of Int. Medicine},
  volume={161},
  number={19},
  pages={2293--2300},
  year={2001},
  publisher={American Medical Association}
}

@article{arima,
  title={ARIMA models},
  author={Shumway, Robert H and Stoffer, David S and Shumway, Robert H and Stoffer, David S},
  journal={Time series analysis and its applications: with R examples},
  pages={75--163},
  year={2017},
  publisher={Springer}
}

@article{bent2021engineering,
  title={Engineering digital biomarkers of interstitial glucose from noninvasive smartwatches},
  author={Bent, Brinnae and Cho, Peter J and Henriquez, Maria and Wittmann, April and Thacker, Connie and Feinglos, Mark and Crowley, Matthew J and Dunn, Jessilyn P},
  journal={NPJ Digital Medicine},
  volume={4},
  number={1},
  pages={89},
  year={2021},
  publisher={Nature Publishing Group UK London}
}

@article{seo2021personalized,
  title={A personalized blood glucose level prediction model with a fine-tuning strategy: A proof-of-concept study},
  author={Seo, Wonju and Park, Sung-Woon and Kim, Namho and Jin, Sang-Man and Park, Sung-Min},
  journal={Computer methods and programs in biomedicine},
  volume={211},
  pages={106424},
  year={2021},
  publisher={Elsevier}
}

@article{langarica2024deep,
  title={Deep learning-based glucose prediction models: a guide for practitioners and a curated dataset for improved diabetes management},
  author={Langarica, Sa{\'u}l and de la Vega, Diego and Cariman, Nawel and Miranda, Mart{\'\i}n and Andrade, David C and N{\'u}{\~n}ez, Felipe and Rodriguez-Fernandez, Maria},
  journal={IEEE Open Journal of Engineering in Medicine and Biology},
  volume={5},
  pages={467--475},
  year={2024},
  publisher={IEEE}
}

@article{lara2025personalized,
  title={Personalized glucose forecasting for people with type 1 diabetes using large language models},
  author={Lara-Abelenda, Francisco J and Chushig-Muzo, David and Peiro-Corbacho, Pablo and W{\"a}gner, Ana M and Granja, Concei{\c{c}}{\~a}o and Soguero-Ruiz, Cristina},
  journal={Computer Methods and Programs in Biomedicine},
  volume={265},
  pages={108737},
  year={2025},
  publisher={Elsevier}
}

@article{langarica2023meta,
  title={A meta-learning approach to personalized blood glucose prediction in type 1 diabetes},
  author={Langarica, Sa{\'u}l and Rodriguez-Fernandez, Maria and N{\'u}{\~n}ez, Felipe and Doyle III, Francis J},
  journal={Control Engineering Practice},
  volume={135},
  pages={105498},
  year={2023},
  publisher={Elsevier}
}

@ARTICLE{9813400,
  author={Zhu, Taiyu and Li, Kezhi and Herrero, Pau and Georgiou, Pantelis},
  journal={IEEE Transactions on Biomedical Engineering}, 
  title={Personalized Blood Glucose Prediction for Type 1 Diabetes Using Evidential Deep Learning and Meta-Learning}, 
  year={2023},
  volume={70},
  number={1},
  pages={193-204},
  doi={10.1109/TBME.2022.3187703}}

@ARTICLE{6157604,
  author={Zecchin, Chiara and Facchinetti, Andrea and Sparacino, Giovanni and De Nicolao, Giuseppe and Cobelli, Claudio},
  journal={IEEE Transactions on Biomedical Engineering}, 
  title={Neural Network Incorporating Meal Information Improves Accuracy of Short-Time Prediction of Glucose Concentration}, 
  year={2012},
  volume={59},
  number={6},
  pages={1550-1560},
  doi={10.1109/TBME.2012.2188893}}

@inproceedings{blood,
  title={Blood glucose level prediction as time-series modeling using sequence-to-sequence neural networks},
  author={Bhimireddy, Ananth and Sinha, Priyanshu and Oluwalade, Bolu and Gichoya, Judy Wawira and Purkayastha, Saptarshi},
  booktitle={KDH@ECAI},
  year={2020},
  organization={CEUR Workshop Proceedings}
}

@article{deep,
  title={Deep multitask learning by stacked long short-term memory for predicting personalized blood glucose concentration},
  author={Shuvo, Md Maruf Hossain and Islam, Syed Kamrul},
  journal={IEEE Journal of Biomedical and Health Informatics},
  volume={27},
  number={3},
  pages={1612--1623},
  year={2023},
  publisher={IEEE}
}

@inproceedings{bg-bert,
  title={Predicting Adverse Events for Patients with Type-1 Diabetes Via Self-Supervised Learning},
  author={Zheng, Xinzhe and Ji, Sijie and Wu, Chenshu},
  booktitle={IEEE Int. Conf. on Acoustics, Speech and Signal Processing (ICASSP)},
  pages={1526--1530},
  year={2024},
  organization={IEEE}
}

@article{deng2021deep,
  title={Deep transfer learning and data augmentation improve glucose levels prediction in type 2 diabetes patients},
  author={Deng, Yixiang and Lu, Lu and Aponte, Laura and Angelidi, Angeliki M and Novak, Vera and Karniadakis, George Em and Mantzoros, Christos S},
  journal={NPJ Digital Medicine},
  volume={4},
  number={1},
  pages={109},
  year={2021},
  publisher={Nature Publishing Group UK London}
}

@article{ohio,
    author = "Marling C., Bunescu R.",
    title = "{The OhioT1DM Dataset for Blood Glucose Level Prediction: Update 2020}",
    journal = "CEUR workshop proceedings",
    vol = "2675",
    pages = "71--74",
    year = "2020"
}

@article{diatrend,
  title={DiaTrend: A dataset from advanced diabetes technology to enable development of novel analytic solutions},
  author={Prioleau, Temiloluwa and Bartolome, Abigail and Comi, Richard and Stanger, Catherine},
  journal={Scientific Data},
  volume={10},
  number={1},
  pages={556},
  year={2023},
  publisher={Nature Publishing Group UK London}
}

@article{butt2023feature,
  title={Feature transformation for efficient blood glucose prediction in type 1 diabetes mellitus patients},
  author={Butt, Hatim and Khosa, Ikramullah and Iftikhar, Muhammad Aksam},
  journal={Diagnostics},
  volume={13},
  number={3},
  pages={340},
  year={2023},
  publisher={MDPI}
}

@article{chawla2002smote,
  title={SMOTE: synthetic minority over-sampling technique},
  author={Chawla, Nitesh V and Bowyer, Kevin W and Hall, Lawrence O and Kegelmeyer, W Philip},
  journal={Journal of artificial intelligence research},
  volume={16},
  pages={321--357},
  year={2002}
}

@inproceedings{shrinkage,
  title={Deep regression tracking with shrinkage loss},
  author={Lu, Xiankai and Ma, Chao and Ni, Bingbing and Yang, Xiaokang and Reid, Ian and Yang, Ming-Hsuan},
  booktitle={Proceedings of the European conference on computer vision (ECCV)},
  pages={353--369},
  year={2018}
}

@inproceedings{rigamonti2024improving,
  title={Improving Detection of Type-1 Diabetes Adverse Events Using GRU Networks},
  author={Rigamonti, Giorgia and Barbato, Mirko Paolo and Marelli, Davide and Napoletano, Paolo},
  booktitle={2024 IEEE 8th Forum on Research and Technologies for Society and Industry Innovation (RTSI)},
  pages={79--84},
  year={2024},
  organization={IEEE}
}

\end{document}